\documentclass[sigconf]{acmart}

\usepackage{enumitem}
\usepackage{graphicx}
\usepackage{multirow}

\AtBeginDocument{%
  }

\copyrightyear{2025}
\acmYear{2025}
\setcopyright{cc}
\setcctype{by-nc-sa}
\acmConference[MM '25]{Proceedings of the 33rd ACM International Conference on Multimedia}{October 27--31, 2025}{Dublin, Ireland}
\acmBooktitle{Proceedings of the 33rd ACM International Conference on Multimedia (MM '25), October 27--31, 2025, Dublin, Ireland}\acmDOI{10.1145/3746027.3758273}
\acmISBN{979-8-4007-2035-2/2025/10}

\begin{document}

\title{Waymo-3DSkelMo: A Multi-Agent 3D Skeletal Motion Dataset for Pedestrian Interaction Modeling in Autonomous Driving}


\author{Guangxun Zhu}
\email{g.zhu.1@research.gla.ac.uk}
\affiliation{%
  \institution{University of Glasgow}
  \city{Glasgow}
  \country{United Kingdom}
}

\author{Shiyu Fan}
\email{s.fan.1@research.gla.ac.uk}
\affiliation{%
  \institution{University of Glasgow}
  \city{Glasgow}
  \country{United Kingdom}
}

\author{Hang Dai}
\email{daihang@whu.edu.cn}
\affiliation{%
  \institution{Wuhan University }
  \city{Wuhan}
  \country{China}
}
\affiliation{%
  \institution{University of Glasgow}
  \city{Glasgow}
  \country{United Kingdom}
}

\author{Edmond S. L. Ho}
\authornote{Corresponding author}
\email{Shu-Lim.Ho@glasgow.ac.uk}
\orcid{0000-0001-5862-106X}
\affiliation{%
  \institution{University of Glasgow}
  \city{Glasgow}
  \country{United Kingdom}
}

\renewcommand{\shortauthors}{Zhu et al.}

\begin{abstract}
Large-scale high-quality 3D motion datasets with multi-person interactions are crucial for data-driven models in autonomous driving to achieve fine-grained pedestrian interaction understanding in dynamic urban environments. However, existing datasets mostly rely on estimating 3D poses from monocular RGB video frames, which suffer from occlusion and lack of temporal continuity, thus resulting in unrealistic and low-quality human motion. In this paper,  we introduce Waymo-3DSkelMo, the first large-scale dataset providing high-quality, temporally coherent 3D skeletal motions with explicit interaction semantics, derived from the Waymo Perception dataset. Our key insight is to utilize 3D human body shape and motion priors to enhance the quality of the 3D pose sequences extracted from the raw LiDRA point clouds. The dataset covers over 14,000 seconds across more than 800 real driving scenarios, including rich interactions among an average of 27 agents per scene (with up to 250 agents in the largest scene). Furthermore, we establish 3D pose forecasting benchmarks under varying pedestrian densities, and the results demonstrate its value as a foundational resource for future research on fine-grained human behavior understanding in complex urban environments. The dataset and code will be available at \url{https://github.com/GuangxunZhu/Waymo-3DSkelMo}%
\end{abstract}


\begin{teaserfigure}
  \includegraphics[width=\textwidth]{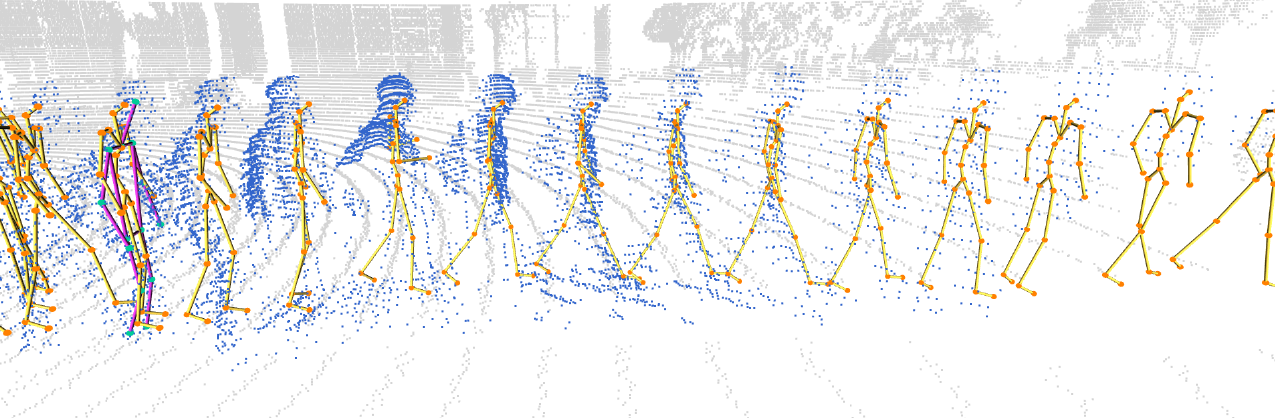}
  \caption{Waymo-3DSkelMo: A high-quality 3D Multi-pedestrian motion dataset created using human motion and shape priors from LiDAR range images in the Waymo perception dataset. 
  (Blue) The point clouds, 
  sampled every 0.5 seconds, of a pedestrian from the LiDAR range images. A 3D body mesh can be estimated from the partial LiDAR point cloud using a 3D human shape prior for each sample.  
  (Purple) The Waymo dataset comes with very sparsely annotated 3D skeletal poses. (Yellow) Based on the skeletal poses extracted from the estimated body meshes, 
  a motion prior is used to enhance the motion quality.}  
  \Description{Waymo-3DSkelMo enables continuous 3D skeletal motion modeling in autonomous driving scenarios, overcoming the sparse single-frame annotation limitations of the original Waymo dataset through human shape and motion priors.}
  \label{fig:waymo_3dskelmo}
\end{teaserfigure}


\maketitle

\section{Introduction}
Understanding interactions between intelligent agents (e.g. pedestrians and vehicles) and their surrounding environment is crucial for autonomous vehicles to achieve accurate perception and safe planning in dynamic urban ecosystems. Recent advances have demonstrated that interaction-aware perception systems produce significant breakthroughs in tasks such as pedestrian trajectory prediction~\cite{SocialCircle} and 3D pose forecasting~\cite{peng2023trajectory,xiao2024multi}. 

The main challenge in interaction modeling lies in capturing the fine-grained dynamics of pedestrian behavior. Pedestrians, as indispensable participants in road environments, need special attention in the development of autonomous driving technologies. 
For example, they actively avoid collisions with other road users (e.g., pedestrians, vehicles, cyclists, etc.), 
while their movements are constrained by roads, traffic signs, and surrounding infrastructure. Inaccurate or coarse modeling can 
lead to incorrect predictions of the future trajectory of pedestrians, which may mislead autonomous vehicles to take actions in response to the situation. 

Most existing works~\cite{azarmi2024pip, Crosato:TIV2023, PedFormer} model interactions between pedestrians using 2D locations, 2D poses, etc. Recently, some studies have attempted to incorporate 3D human poses to better capture motion dynamics~\cite{Distillation, saadatnejad2023social}. Compared to 2D data, 3D human pose provides richer spatial structure and depth information, offering a stronger foundation for fine-grained interaction modeling. However, due to the lack of fully annotated 3D data in existing public datasets on pedestrian interactions (see Table~\ref{tab:dataset_stats}), researchers have been collecting multi-pedestrian 3D skeletal motion data by applying state-of-the-art 3D pose estimation from RGB image frames \cite{Jeong:CVPR2024,Distillation}, such as JRDB-GlobMultiPose
(JRDB-GMP)~\cite{Jeong:CVPR2024} and MuPoTS-3D~\cite{MuPoTs-3D}. However, that resulted in poor motion quality due to the absence of temporal information from frame-based pose estimation and heavy occlusion when handling scenes with multiple pedestrians interacting with each other. \citet{Wang:NeurIPS2021} explored combining single-person and two-person motions as synthetic 3-person interaction using data from the high-quality CMU-MOCAP dataset~\cite{mocapdata} which was collected using optical Motion Capture (MOCAP) system. However, this results in a synthetic dataset with low diversity. There is also a stream of research that focuses on
estimating 3D poses directly from noisy, sparse, and incomplete LiDAR point clouds. 
This process can introduce a lot of noise and lacks prior knowledge about the human body and motion~\cite{ye2024lpformer}. Although some methods consider motion consistency via neighborhood enhancement for short-term coherence \cite{zhang2024neighborhood}, their modeling remains spatially local without considering long-term or global temporal information. 



To mitigate the challenges of data scarcity and enhance data quality, we present Waymo-3DSkelMo (Figure~\ref{fig:waymo_3dskelmo}). To our best knowledge, this is the first large-scale dataset 
to provide high-quality 
3D skeletal motions with explicit interaction semantics, all derived from raw LiDAR range images in the Waymo Open Dataset Perception Benchmark (hereafter referred to as Waymo)~\cite{Waymo:Sun_2020_CVPR} which were captured from real-world outdoor scenarios. 
Our dataset overcomes the aforementioned issues by using human body shape 
priors through SMPL~\cite{loper2023smpl} mesh recovery and enhancing the naturalness of the 3D motion using human motion priors through Neural Motion Field (NeMF)~\cite{NeMF}. 
Waymo-3DSkelMo thus bridges the gap between raw sensor data and interaction decoding requirements. Specifically, our dataset covers more than 14,000 seconds of recordings from over 800 real driving scenarios, capturing interactions among an average of 27 agents per scene and reaching 250 agents in the most crowded scenarios, providing abundant multi-agent interaction patterns for robust modeling. Moreover, our dataset is spatially and temporally synchronized with the original Waymo dataset, allowing seamless cross-modal integration with existing LiDAR/camera sensor data and annotations. Finally, to demonstrate the potential application of our dataset, we establish rigorous benchmarks for 3D pose forecasting, with protocols for short-term (1s)
prediction across scenes with varying numbers of pedestrians.


Our contributions can be summarized as follows:

\begin{itemize}
  \item We propose a new pipeline to adopt SMPL-based mesh recovery to acquire higher-quality 3D pose from raw LiDAR data and further enhance the motion quality using a motion prior trained with millions of frames.  
  \item We release the first large‑scale autonomous driving dataset featuring continuous, occlusion‑robust 3D skeletal motions with explicit interaction semantics, fully aligned with the Waymo dataset.
  \item We benchmark 3D pose forecasting performance across scenes with varying pedestrian densities to facilitate in-depth analysis of interaction-aware motion prediction and serve as a foundation for future research on fine-grained human behavior understanding in complex driving environments.

\end{itemize}
\noindent

\begin{figure}
  \centering
  \includegraphics[width=\columnwidth]{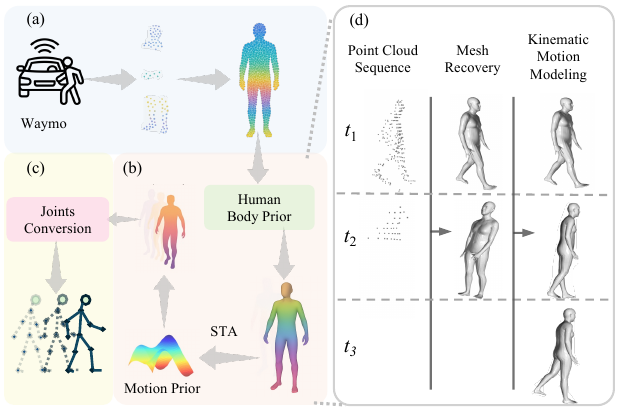}
  \caption{Overview of our 
  pipeline. (a) 
  Point clouds are first extracted from the range images of all five Waymo LiDAR sensors, then transformed into a world coordinate system and fused into a unified point cloud representation. (b) 
  Mesh recovery is performed on all point clouds using a human‐body prior, followed by motion generation via a motion prior. (c) 
  Regressing SMPL parameters to 
  skeletal motions. (d) An example of different quality of point cloud and 3D pose.}
  \Description{Overview of the 3D Skeletal Motion Dataset Generation Pipeline.}
  \label{fig:pipeline}
\end{figure}


\begin{table*}[htbp]
\centering
\caption{Comparison of statistics between the newly proposed Waymo-3DSkelMo and existing human pose forecasting datasets
}
\label{tab:dataset_stats}
\begin{tabular}{@{}p{1.1in}p{1in}p{1in}p{1in}p{1in}p{1in}@{}}
\toprule
\textbf{} & \textbf{Waymo-3DSkelMo (Ours)} & \textbf{Waymo}~\cite{Waymo:Sun_2020_CVPR} & \textbf{MuPoTS-3D}~\cite{MuPoTs-3D} & \textbf{CMU-Mocap
(UMPM)}~\cite{peng2023trajectory} & \textbf{JRDB-GMP}~\cite{Jeong:CVPR2024} \\
\midrule
Scenes w/pedestrains   & \textbf{837}        & \textbf{837}       & 20    & Partially synthetic     & 27\\
Duration \textit{(s)}              & 14419 & \textbf{14421}     &   267 & 4,000     & 1863 \\
avg. pedestrians \#    & 27.1      & \textbf{27.6}       & 3  & 3 & 6.8 \\
max. pedestrians \#     & 250    & \textbf{253}      & 3     & 3     & 24 \\
avg. displacement \textit{(m)}     & \textbf{10.41}          &  10.33         & 0.55       & 0.63         &  0.76   \\
3D poses \# \textit{(@10 fps)}  & \textbf{2,438,145}& 9,976        &  8,010   & 120,000    & 400,000\\
Collection method \textit{(PE=Pose Estimation)}      & LiDAR PE + motion enhancement & Manual annotation & RGB Image PE & Optical MOCAP sequences + mixing & RGB Image PE \\
\bottomrule
\end{tabular}
\end{table*}



\section{Description of the dataset} \label{sec:datasetDesc}

Our dataset contains over 14,000 seconds of pedestrian 3D skeletal motion sequences across 837 distinct real-world scenes, with an average of 27 agents interacting per scene.
 The number of pedestrians per scene ranges from 1 to over 250. The raw data are sourced from the training and validation sets of the Waymo dataset. Please refer to Section~\ref{sec:waymo} for the details of the original Waymo dataset. 

Table~\ref{tab:dataset_stats} presents a quantitative comparison between the combined training and validation splits of the original Waymo dataset 
and our proposed Waymo-3DSkelMo. Note that test sets are excluded due to licensing restrictions. 
Both datasets share the same 837 scenes containing pedestrians. While Waymo-3DSkelMo shows a slightly lower total duration (14,419 ~\textit{s} vs. 14,421 ~\textit{s}) and number of pedestrians per scene (27.1 vs. 27.6 average; 250 vs. 253 maximum) compared to the original Waymo dataset, this is primarily due to missing point cloud data in the original dataset that covers the full duration of some pedestrian motion sequences. In such cases, generating reliable and temporally consistent 3D annotations is infeasible, and those sequences are therefore excluded from our version, 
leading to a higher average displacement (10.41 ~\textit{m}  vs. 10.33 ~\textit{m}). 
In particular, the original Waymo dataset provides only about 10,000 manually annotated 3D poses for pedestrians across all scenes, resulting in sparse and isolated ground-truth labels. On the other hand, Waymo-3DSkelMo offers dense 3D skeletal motion annotations for all pedestrians at every time step, totaling 2,438,145 frame-level annotations \textit{(@10 fps)} generated using human body and motion priors. This dense supervision enables fine-grained temporal modeling and mitigates the sparsity and inconsistency present in the original keypoint annotations.

Moreover, both datasets contain highly interactive scenes with up to 250 pedestrians appearing in a single scene, which makes Waymo-3DSkelMo particularly valuable for modeling large-scale multi-agent interaction dynamics in autonomous driving contexts.

Compared to MuPoTS-3D~\cite{MuPoTs-3D}, CMU-Mocap (UMPM)~\cite{peng2023trajectory}, and JRDB-GMP~\cite{Jeong:CVPR2024}, which are also densely annotated, Waymo-3DSkelMo offers significantly larger scale, richer interactions, and more dynamic motion. While the three datasets contain on average only 3 to 6.8 pedestrians per scene with a duration of up to 4,000 seconds, Waymo-3DSkelMo spans over 14,000 seconds, includes up to 250 pedestrians per scene (27.1 on average), and achieves a higher average displacement of 10.41 meters, reflecting longer trajectories which contain more interactions among the pedestrians.

In addition, our dataset provides detailed 3D pedestrian motion annotations in two main formats: keypoint-based data and SMPL mesh data. The keypoint format includes 3D coordinates of human skeletal joints, while the SMPL format offers a comprehensive parametric human mesh capturing detailed body shape and pose. To accommodate different research needs, we supply the dataset at two frame rates: 30 fps and 10 fps. The 10 fps version is temporally synchronized with the original Waymo dataset, with aligned timestamps, making it suitable for use as ground truth annotations. 

\begin{figure}
  \centering
  \includegraphics[width=\columnwidth]{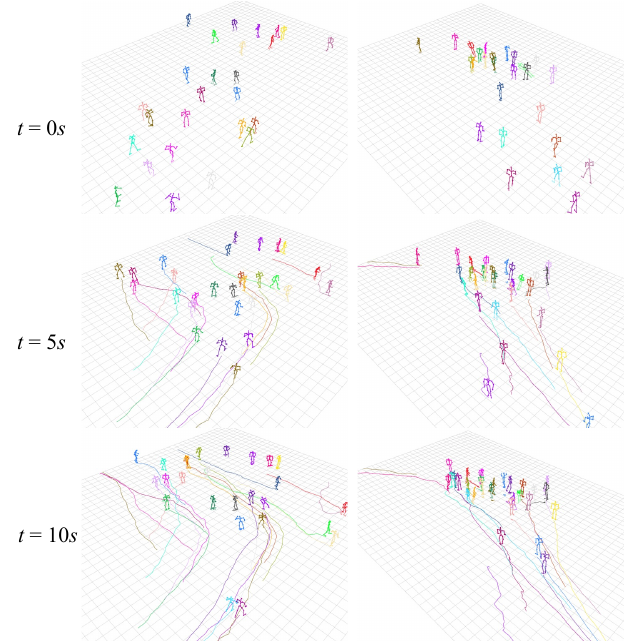}
  \caption{Example scenes from the Waymo-3DSkelMo dataset, illustrating its multi-agent interaction nature.
  }
  \Description{}
  \label{fig:dataset_example}
\end{figure}

\section{Dataset Generation Process}

An overview of our 3D skeletal motion dataset generation is illustrated in Figure~\ref{fig:pipeline}. Our dataset described in Section~\ref{sec:datasetDesc} is generated from the raw LiDAR range images provided in the Waymo dataset~\cite{Waymo:Sun_2020_CVPR}. First, point clouds are extracted from the range images captured from all five Waymo LiDAR sensors, reprojected into a common world‐coordinate frame via extrinsic calibration, and merged into a unified 3D point cloud representation. We then utilise LiDAR-HMR~\cite{fan2023lidar} to recover 3D Human Mesh based on a 3D human shape prior from the raw LiDAR data. The output of LiDAR-HMR contains noisy and missing data, and we mitigate these issues by spatiotemporal alignment. We further take advantage of using 3D human skeletal motion prior to improve the motion quality of the refined output using the Neural Motion Field (NeMF)~\cite{NeMF}. The details are explained in the following subsections.


\subsection{Waymo Perception Dataset}
\label{sec:waymo}
The Waymo Open Dataset Perception Benchmark~\cite{Waymo:Sun_2020_CVPR} 
introduced human keypoint annotations in version 1.3.2, which includes LiDAR range images along with corresponding camera images. For our experiments, we use version 2.0 for training and validation. This version provides annotations of 14 keypoints (as a pose, from nose to ankle) per pedestrian. The training set contains 8,125 annotated poses, while the validation set includes 1,873. However, the raw LiDAR point clouds often suffer from sparsity and noise, especially at longer ranges, resulting in missing or unevenly distributed 3D points around occluded limbs and small body parts. These quality issues make it difficult to obtain reliable 3D poses from the raw LiDAR data.


\subsection{3D Human Mesh Recovery}

To accurately capture human body shape and pose from sparse, noisy, and incomplete LiDAR data, we adopt LiDAR-HMR~\cite{fan2023lidar}, a state-of-the-art method for 3D human mesh recovery. LiDAR-HMR leverages a parametric human body model (SMPL) as a prior to reconstruct detailed 3D meshes from raw LiDAR range images, allowing robust recovery even under occlusions and partial observations.

As shown in Figure~\ref{fig:pipeline} (d), in our pipeline, LiDAR-HMR processes LiDAR observations of each pedestrian frame-by-frame, treating each pedestrian at every timestamp as an individual instance. We perform human mesh recovery on all point clouds, even if a point cloud contains only a single point. For high-quality and relatively complete point clouds (e.g., at \( t_1 \) in Figure~\ref{fig:pipeline}), LiDAR-HMR can generate accurate and reliable meshes. However, when point clouds are incomplete or noisy, the recovered meshes often suffer from issues such as inconsistent orientations, incorrect heights, and poses that are not consistent with the point cloud data or temporal motion patterns  (e.g., at \( t_2 \) in Figure~\ref{fig:pipeline}). Moreover, in the absence of point cloud data, mesh recovery is not feasible  (e.g., at \( t_3 \) in Figure~\ref{fig:pipeline}).



\subsection{Spatiotemporal Alignment} \label{sec:SAlign}
To achieve more coherent motion, we perform spatiotemporal alignment (STA) through temporal interpolation and Frenet frame~\cite{do2016differential} transformation. Specifically, we first apply spherical linear interpolation (slerp) for joint rotations and linear interpolation for translations to fill missing frames in pedestrian sequences. The data is then upsampled to $30\,$ fps to match the next processing step. To resolve inconsistent global orientations, we adopt the Frenet frame, which constructs a local coordinate system along motion trajectories using tangent ($\mathbf{T}$), normal ($\mathbf{N}$), and binormal ($\mathbf{B}$) vectors, ensuring stable orientation representation relative to movement direction. Together, these spatiotemporal alignment steps yield smoother, consistently oriented 3D motion that is better suited for subsequent processing. The improvements are highlighted in Table~\ref{tab:generation_results}.


\subsection{Neural Motion Fields}
To further overcome temporal inconsistencies and missing frames in the recovered 3D meshes, we utilize Neural Motion Fields (NeMF)~\cite{NeMF}. NeMF offers a continuous and smooth representation of kinematic motion, enabling the interpolation of missing data and enforcing realistic motion dynamics. Since NeMF is pre-trained using the AMASS~\cite{AMASS} dataset which contains over 40 hours of high-quality 3D motion data (>11,000 motion sequences) collected from over 300 subjects, this enables us to use NeMF as a generic motion prior to enhance the quality of the motion sequences.




Given the interpolated poses, we first encode the 3D skeletal pose sequence into NeMF motion latent representation ($z_l$ for the \textit{local} pose relative to the root joint and $z_g$ for \textit{global} translation of the root joint) using the pre-trained NeMF encoder and optimize it based on the energy function: 
\begin{equation}
z_{l}^{*}, z_{g}^{*} = \underset{z_l, z_g}{\mathrm{argmin}} \sum_{\mathcal{T}}{\lambda_{rot}\mathcal{L}_{rot}+\lambda_{ori}\mathcal{L}_{ori}+\lambda_{pos}\mathcal{L}_{pos}+\lambda_{trans}\mathcal{L}_{trans}}    
\end{equation}
where $\mathcal{L}_{rot}$ and $\mathcal{L}_{pos}$ are the $L_1$ losses on the 6D rotations \cite{6D_rot} and the 3D local positions of all joints, 
$\mathcal{L}_{ori}$ and $\mathcal{L}_{trans}$ are the $L_1$ losses on the orientation and translation of the root joint, respectively, and $\mathcal{T}$ is a set of all interpolated frames from the spatiotemporal alignment (Section~\ref{sec:SAlign}). The $L_1$ losses are calculated between the interpolated 3D skeletal motion sequence and the motion decoded from the intermediate NeMF representation (i.e. $z_{l}^{*}$ and $z_{g}^{*}$). Furthermore, the loss terms for keyframes (those with ground-truth annotations in the original Waymo dataset) are weighted $10\times$ higher than those for other frames, in order to mitigate error accumulation and obtain motion sequences that better reflect realistic human movements. By this, the resultant motion will be as similar to the interpolated motion as possible while preserving the naturalness as represented in the NeMF. We empirically found $\lambda_{rot}=\lambda_{ori} = 1$ and $\lambda_{pos}=\lambda_{trans}=10$ 
yield the best outcomes.

\subsection{Generation results and evaluation}

To validate the effectiveness of our pipeline, we conduct comprehensive comparisons across four paradigms:

\begin{itemize}
    \item \textbf{LiDAR-HMR Raw}: Direct outputs from the LiDAR-HMR without refinement.
    \item \textbf{Linear Interpolation}: Frame-wise linear filling of missing poses within LiDR-HMR output. 
    \item \textbf{CondMDI-based}: Motion inbetweening via the state-of-the-art CondMDI model~\cite{cohan2024flexible}, using LiDAR-HMR outputs as keyframes to complete the missing frames.
    \item \textbf{Ours}: Full pipeline with NeMF‑based motion reconstruction.
\end{itemize}
\noindent 


To quantitatively evaluate motion quality, we compare the motion generated using the aforementioned methods by four established metrics: Fréchet Inception Distance (FID) \cite{NeMF, cohan2024flexible,xu2025multi}, which measures the distance between the distribution of generated motions and that of real motions from the HumanML3D dataset \cite{guo2022generating} to indicate overall realism; Jittering \cite{karunratanakul2023guided,xu2025multi}, computed as the average $L_2$ norm of joint acceleration vectors across all joints and frames, to quantify frame‑to‑frame smoothness; Mean Per Joint Position Error (MPJPE) \cite{ye2024lpformer,fan2023lidar}, defined as the mean Euclidean distance between the generated joint position and the ground‑truth position at frames with annotated corresponding 3D keypoints in Waymo; and Foot Skating Ratio (FS) \cite{cohan2024flexible, NeMF, xu2025multi}, the percentage of frames in which foot‑ground penetration exceeds $2\,$ cm, to assess the physical plausibility of the motion.




\begin{table*}[htbp]
\centering
\caption{Quantitative comparison of motion generation methods with and without Frenet‑frame alignment. Metrics marked with $\downarrow$ indicate that lower values are better. Within each setting (with/without Frenet), the best result for each metric is highlighted in bold.}
\Description{Quantitative comparison of motion generation methods with and without Frenet‑frame alignment.}
\label{tab:generation_results}
\begin{tabular}{@{}l l c c c c@{}}
\toprule
\textbf{Setting} & \textbf{Method} & \textbf{FID} $\downarrow$ & \textbf{Foot Skating} $\downarrow$ & \textbf{MPJPE (m) $\downarrow$} & \textbf{Jittering (m/f$^2$)} $\downarrow$ \\
\midrule
\multirow{4}{*}{\textbf{w/o Frenet}} 
  & LiDAR-HMR Raw         & 15.5387          & 0.032430          & \textbf{0.0869}   & 0.315535          \\
  & Linear Interpolation  & 19.1160          & 0.032525          & \textbf{0.0869}   & 0.250194          \\
  & CondMDI-based         & 14.6586          & 0.051199          & 0.8241            & 0.250889          \\
  & \textbf{NeMF-based (Ours)}         
  & \textbf{12.2387} & \textbf{0.030393} & 0.1391            & \textbf{0.047769} \\
  
\midrule
\multirow{4}{*}{\textbf{Frenet}}     
  & LiDAR-HMR Raw         & 15.8628          & 0.033833          & \textbf{0.1263}   & 0.314766          \\
  & Linear Interpolation  & 19.9499          & \textbf{0.033603}          & \textbf{0.1263}   & 0.249554          \\
  & CondMDI-based         & 16.7842          & 0.038380          & 0.8870            & 0.228974          \\
  & \textbf{NeMF-based (Ours)}         
    & \textbf{10.4501} & 0.034975 & 0.1705            & \textbf{0.047220} \\

\bottomrule
\end{tabular}
\end{table*}

Quantitative results are presented in Table~\ref{tab:generation_results}. Our method achieves the lowest FID in both settings (12.2387 w/o Frenet, 10.4501 w/ Frenet), indicating that its motion distribution resembles the real pedestrian data the most closely. 
The improvement in FID achieved through Frenet alignment indicates that resolving inconsistent global orientations leads to motion sequences that more accurately reflect realistic human movement patterns.
It also attains the 
lowest foot-skating ratios (0.030393 w/o Frenet, 0.034975 w/ Frenet), reflecting superior contact realism, though the slight degradation under Frenet alignment results from correcting systematic orientation drift. The lowest joint‑acceleration jitter (0.047769 w/o Frenet, 0.047220 w/ Frenet) further demonstrates smoother trajectories, particularly after applying Frenet‑frame corrections. 

Although raw LiDAR‑HMR and linear interpolation preserve joint positions most accurately (0.0869 w/o Frenet, 0.1263 w/ Frenet), they suffer from poor pose expressiveness and temporal stability. In contrast, slightly increase in pose position error in our pipeline (0.1391 w/o Frenet, 0.1705 w/ Frenet) is outweighed by substantial gains in  
temporal coherence, and anatomical plausibility—benefits that are further enhanced by orientation correction. CondMDI, on the other hand, underperforms across all metrics, as its reliance on keyframe quality and distribution propagates noise and leads to inconsistent motion between frames.

Additionally, Figure~\ref{fig:dataset_example} visualizes some scenes from the  Waymo-3DSkelMo dataset. Even in high-density scenarios (up to 30 pedestrians), the motion remains remarkably smooth and collision-free. The dataset spans pedestrian motions of both long and short distances and showcases extensive agent-to-agent interaction in terms of both trajectories and local poses.


\begin{table*}[htbp]
\centering
\caption{Results of JPE, APE and FDE (in mm) under different number of persons settings. We compare short-term predictions using TBIFormer across varying person interactions.}
\Description{Results of JPE, APE and FDE (in mm) under different number of persons settings. We compare short-term predictions using TBIFormer across varying person interactions.}
\label{tab:multi_person_metrics}
\begin{tabular}{l c c c c c c c c c c c c c c c c}
\toprule
& \multicolumn{4}{c}{\textbf{2 persons}} 
& \multicolumn{4}{c}{\textbf{3 persons}} 
& \multicolumn{4}{c}{\textbf{4 persons}} 
& \multicolumn{4}{c}{\textbf{5 persons}} \\
\cmidrule(lr){2-5} \cmidrule(lr){6-9} \cmidrule(lr){10-13} \cmidrule(lr){14-17}
\textbf{} 
& 0.2s & 0.6s & 1.0s & Overall 
& 0.2s & 0.6s & 1.0s & Overall 
& 0.2s & 0.6s & 1.0s & Overall 
& 0.2s & 0.6s & 1.0s & Overall \\
\midrule
JPE & 110 & 366 & 504 & 326 & 89 & 190 & 237 & 172 & 98 & 200 & 276 & 191  & 113 & 261 & 368 & 247 \\
APE &  72 & 130 & 124 & 109 & 78 & 141 & 157 & 125 & 83 & 137 & 152 & 124  &  90 & 149 & 153 & 131 \\
FDE &  80 & 313 & 469 & 287 &  51 & 120 & 159 &110 & 56 & 143 & 221 & 140  &  67 & 199 & 316 & 194 \\

\bottomrule
\end{tabular}
\end{table*}

\section{Benchmarking on the dataset}
To validate the effectiveness of our dataset, we benchmark it using TBIFormer~\cite{peng2023trajectory}, an interaction-aware 3D pose forecasting model. 

\subsection{Dataset}
 We follow the original training and validation splits of Waymo, specifically selecting scenes containing 2 to 5 pedestrians for benchmarking, as these represent typical urban interaction scenarios. To align with TBIFormer's temporal resolution, the NeMF‑based generated sequences, which are originally at 30 fps, are downsampled to 25 fps. For short-term prediction, we use 2 seconds (50 frames) of historical observation to predict 1 second (25 frames) of future poses. Overlapping segments with consistent pedestrian counts are then identified through timestamp alignment. Subsequently, a sliding window with a duration of 3 seconds (75 frames) and a step size of 50 frames is applied to extract clips. In addition, data augmentation techniques are employed, including global rotations around the vertical axis in discrete steps of 10°, Gaussian noise ($\sigma = 0.01\,\mathrm{m}$
) added to joint positions, and random spatial scaling between 0.9 and 1.1, to enhance the diversity of the dataset.

\subsection{Implementation details}
We implement our framework in PyTorch, and the experiments are performed on a single Nvidia GeForce RTX 4090 GPU. We train our model for 50 epochs using the ADAM optimizer with a batch size of 32, a learning rate of 0.00002, and a dropout rate of 0.2. All other settings remain consistent with TBIFormer. Notably, separate models are trained for different numbers of pedestrians to better accommodate the variations in interaction distributions.

\subsection{Evaluation Metrics}

We adopt the same evaluation metrics as TBIformer\cite{peng2023trajectory}: the Joint Position Error (JPE), Aligned Mean Per Joint Position Error (APE) and Final Displacement Error (FDE). JPE measures the average per‑joint Euclidean distance between predicted and ground‑truth poses across all pedestrians, while APE further removes global translation by aligning root joints before computing JPE. FDE assesses the accuracy of predicted root positions at the final time step.

\subsection{Results}
Table \ref{tab:multi_person_metrics} presents the short-horizon 
prediction errors of TBIFormer under different numbers of persons, evaluated by 
JPE, APE, and FDE at 0.2s, 0.6s, 1.0s time horizons, as well as overall average errors.

As expected, all three metrics increase with prediction horizon, reflecting growing uncertainty in longer‐term forecasting. Notably, settings with three and four pedestrians yield the lowest overall JPE (172 mm and 191 mm, respectively) and FDE (110 mm and 140 mm, respectively), indicating that modeling moderate levels of interactions between pedestrians enhances predictive accuracy. In contrast, both overly sparse and overly dense pedestrian configurations present challenges. However, the APE is lowest (109 mm) in the two-person case, which indicates that adding more pedestrian interactions can actually introduce extra noise into individual joint predictions, raising the absolute error when three to five people are involved.

In \cite{peng2023trajectory}, the pose forecasting performance on the CMU-Mocap~\cite{mocapdata} (merged with UMPM~\cite{UMPM}) (3 persons) and MuPoTS-3D~\cite{MuPoTs-3D} (2 to 3 persons) datasets was reported. On CMU-Mocap (UMPM), TBIFormer achieved an overall JPE = 107, APE = 76 and FDE = 74, while achieving an overall JPE = 195, APE = 121 and FDE = 63 on MuPoTS-3D. Note that the CMU-Mocap (UMPM) dataset contains a large amount of synthetic 3-person data by mixing single-person and two-person walking motions together as in \cite{Wang:NeurIPS2021}, which results in a low diversity dataset. It can be seen that TBIFormer performed particularly well on this less-challenging dataset. The performance of TBIFormer on MuPoTS-3D and our dataset is similar in overall JPE and APE, while there is a bigger performance gap in FDE. 
The results show that TBIFormer's performance on our dataset is comparable to other datasets, and our work will provide the community with a more challenging dataset for future research. 

It is important to note that our tests only consider interactions between pedestrians and leave out cars, bicycles, and other static or moving objects in the scene. Those missing factors can affect the walking paths of pedestrians which introduce variation and degrade predictive performance under certain conditions.

\section{Conclusion}
In this work, we present Waymo-3DSkelMo, the first large-scale dataset enabling fine-grained interaction modeling through high-quality 3D skeletal motions in real-world driving scenarios. By integrating human shape priors and neural motion field, our dataset provides occlusion-robust and temporally coherent 3D skeletal motions with explicit interaction semantics while maintaining full spatiotemporal alignment with Waymo dataset. Comprising 
14,000 seconds of recordings across 800+ diverse scenarios (1-250+ agents/scene), it establishes a foundational resource validated through 3D pose forecasting benchmarks for modeling complex multi-pedestrian behaviors. Future enhancements will incorporate advanced models for point cloud alignment~\cite{Hu:TIM2023} for pose estimation as well as evaluating the impact on the performance of using different 3D body shape and motion priors. 

\section{Licensing and Access} \label{sec:license}
The enhanced 3D skeletal motion sequences are publicly available for non-commercial use, which also aligns with the Waymo Open Dataset~\cite{Waymo:Sun_2020_CVPR}. Similarly, we will align with the Waymo dataset on limiting the model architectures for non-commercial distribution only, which means the user may publish, in whole or in part, trained model architectures, including weights and biases, developed using our dataset for non-commercial purposes.

We will also open-source the Python implementation for the motion enhancement. All material, including code and dataset, will be made available under Creative Commons BY-NC-SA 4.0 license.


\section{Reproducibility and Supplementary Materials}
As stated in Section~\ref{sec:license}, the code for motion enhancement will be made available as an open-source project. Instructions will be given for accessing the other models and resources used in this project to reproduce the experimental results and the dataset. 

\section{Ethical Considerations and Privacy}
The input data source of our proposed pipeline is fully anonymized. Therefore, there are no ethical considerations or privacy issues.

\begin{acks}
Guangxun Zhu is supported by the funding from the China Scholarship Council (CSC).
\end{acks}


\bibliographystyle{ACM-Reference-Format}
\bibliography{paper}


\end{document}